# Leveraging tensor kernels to reduce objective function mismatch in deep clustering

Daniel J. Trosten [a,*,1], Sigurd Løkse [a,1], Robert Jenssen [b,c,d,2], Michael Kampffmeyer [b,d,2]

[a] NORCE Norwegian Research Centre. Siva Innovasjonssenter, Sykehusveien 21, 9019 Tromsø, Norway
[b] Department of Physics and Technology, UiT The Arctic University of Norway, Hansine Hansens veg 18, 9019 Tromsø, Norway
[c] Department of Computer Science, University of Copenhagen, Universitetsparken 1, Copenhagen 2100, Denmark
[d] Norwegian Computing Center, Gaustadalleen 23a, 0373 Oslo, Norway

ABSTRACT

Objective Function Mismatch (OFM) occurs when the optimization of one objective has a negative impact on the optimization of another objective. In this work we study OFM in deep clustering, and find that the popular autoencoder-based approach to deep clustering can lead to both reduced clustering performance, and a significant amount of OFM between the reconstruction and clustering objectives. To reduce the mismatch, while maintaining the structure-preserving property of an auxiliary objective, we propose a set of new auxiliary objectives for deep clustering, referred to as the Unsupervised Companion Objectives (UCOs). The UCOs rely on a kernel function to formulate a clustering objective on intermediate representations in the network. Generally, intermediate representations can include other dimensions, for instance spatial or temporal, in addition to the feature dimension. We therefore argue that the naïve approach of vectorizing and applying a vector kernel is suboptimal for such representations, as it ignores the information contained in the other dimensions. To address this drawback, we equip the UCOs with structure-exploiting tensor kernels, designed for tensors of arbitrary rank. The UCOs can thus be adapted to a broad class of network architectures. We also propose a novel, regression-based measure of OFM, allowing us to accurately quantify the amount of OFM observed during training. Our experiments show that the OFM between the UCOs and the main clustering objective is lower, compared to a similar autoencoder-based model. Further, we illustrate that the UCOs improve the clustering performance of the model, in contrast to the autoencoder-based approach. The code for our experiments is available at https://github.com/danieltrosten/tk-uco.

## 1. Introduction

Clustering is a long-standing problem in machine learning research. The recent success of deep learning [1] has given rise to *deep clustering* – a subfield of deep learning where deep neural networks are trained with unsupervised loss functions designed for clustering.

Many of the loss functions created for deep clustering attempt to discover and strengthen clusters in the representations produced by the neural network. However, the minimization of these objectives can often result in trivial solutions, where the network maps all inputs to a constant representation, forcing all points to be assigned to the same cluster. This behavior is an extreme case of a more general problem in deep clustering, where minimizing the clustering loss results in an embedding which does not respect the local similarity structure of the input space [2].

To address the similarity preservation problem, several recent deep clustering models employ deep neural networks that have been pre-trained as autoencoders [2–4]. In the majority of these models, clustering is performed in the autoencoder's latent space. The model is fine-tuned by minimizing either the clustering loss alone, or both the clustering loss, and the autoencoder's reconstruction loss. Guo et al. [2] argue that fine-tuning with both the reconstruction loss and clustering loss leads to improved preservation of local similarities in autoencoder-based deep clustering models. They retain the reconstruction loss during fine-tuning, and illustrate that it leads to improved

* Corresponding author.
*E-mail addresses:* dtro@norceresearch.no (D.J. Trosten), sigl@norceresearch.no (S. Løkse), robert.jenssen@uit.no (R. Jenssen), michael.c.kampffmeyer@uit.no (M. Kampffmeyer).
[1] Work done while at the UiT Machine Learning Group, machine-learning.uit.no.
[2] RJ and MK are with the UiT Machine Learning Group, and the Visual Intelligence Centre, visual-intelligence.no. RJ is with the Pioneer Centre for AI, aicentre.dk.

clustering performance for the well known Deep Embedded Clustering (DEC) model [3].

However, clustering and reconstruction are fundamentally different tasks, and thus they require the representations produced by the network to encode different types of information. When analyzing images for instance, successful reconstruction requires the representation to encode the position, color, and fine-grained details of objects in the images. Successful clustering on the other hand, often only depends on higher level information from the image, such as the presence of certain objects or features.

The above properties makes deep clustering models prone to suffer from *Objective Function Mismatch* – a phenomenon where the optimization of an auxiliary objective (*e.g.* reconstruction) has a negative impact on the optimization of the main objective (*e.g.* clustering). OFM has previously been studied in deep clustering by Mrabah et al. [5,6], where they propose Feature Drift (FD) as a way to measure OFM between main and auxiliary objectives.

However, we identify several drawbacks with using FD to measure OFM. The drawbacks are mainly related to the fact that FD can be trivially eliminated by making the auxiliary loss and main loss proportional to each other. Consequently, we argue that reduced FD does not necessarily correspond to improved clustering performance.

Aiming to address the drawbacks of FD, we propose a new, regression-based measure of OFM, which more directly measures the relationship between the auxiliary loss and clustering performance. Moreover, we develop a set of new auxiliary objectives for deep clustering. Our *Unsupervised Companion Objectives* (UCOs) are specifically designed for clustering, in order to reduce the OFM between them and the main clustering objective. In addition, UCOs help preserve the local similarity structure by encouraging a consistent cluster structure throughout the network.

The UCOs rely on a kernel function to measure the similarity between intermediate representations in the network. In general, these intermediate representations may have several dimensions, such as spatial or temporal, in addition to the feature dimension — meaning that they are naturally represented as tensors, rather than vectors. Here, we argue that naïve kernels based on vectorization of tensors do not take this structural (*e.g.* spatial or temporal) information into account. To address these drawbacks, we utilize tensor kernels [7] to measure similarities in the proposed UCOs. The tensor kernels are *structure-exploiting*, meaning that they are able to exploit both structural information and feature information.

We use convolutional neural networks (CNNs) and image data, as well as multi-layer perceptrons (MLPs) and vector data in this work. However, we emphasize that the tensor kernels make the UCOs compatible with a wide range of network architectures and data types, as long as they produce tensorial intermediate representations. Furthermore, the UCOs only depend on the intermediate representations through the kernel function — allowing them to be adapted to all architectures for which a kernel can be specified for the intermediate representations.

In our experiments, we demonstrate that adding the UCOs to different deep clustering models both improves the overall clustering performance, and leads to lower OFM when compared to an analogous autoencoder-based model.

A preliminary version of this paper was published in [8]. Here, we extend several aspects of our previous work:

1. We introduce a new regression-based measure of OFM.
2. We argue that the vectorization-based kernel used in [8] does not take the structural information of intermediate representations into account. To address this issue, we propose an improved version of the UCOs that incorporate structural information by using tensor kernels [7]. This allows us to formulate a structure-exploiting kernel on the intermediate representations of a wide range of deep neural networks, without resolving to suboptimal heuristics, such as vectorization.
3. We conduct an extensive evaluation of the proposed methodology, with several additional experiments. In contrast to [8], which only included image datasets, we include two extra vector-based datasets — illustrating that the UCOs improve clustering performance on these datasets as well. In addition, we analyze the performance of the UCOs across different kernels and weighting strategies, and provide a thorough experimental analysis of the OFM observed for the different models. Finally, we attach the UCOs to additional base models, illustrating that the UCOs work well with these models as well.
4. We provide new insight into the observed increase in OFM during training with the UCOs.
5. We include an extensive overview of related work, placing our work in the current context of deep clustering.

The rest of the paper is structured as follows: Section 2 gives an overview of related work on deep clustering with and without autoencoders, as well as previous work on OFM in different areas of deep learning. In Section 3 we present the proposed UCOs, and in Section 4 we motivate and introduce the tensor kernels, and show how they are adapted to the UCOs. Section 5 discusses the drawbacks of using Feature Drift [5,6] as a measure of OFM in deep clustering, and presents an alternative measure that is better suited to measure the impact of OFM on clustering performance. Section 6 provides the details and results of our experimental evaluation. Section 7 discusses limitations of the proposed methodology, before some concluding remarks are given in Section 8.

## 2. Related work

### 2.1. Deep clustering with autoencoders

Autoencoders are widely used building blocks in deep clustering models. A straightforward approach to integrating autoencoders in a deep clustering model is to use the autoencoder to pre-train the network, and then fine-tune it using only the clustering loss [3,9,10]. Alternatively, there are methods that retain the reconstruction loss during fine-tuning – either by minimizing it jointly with the clustering loss [2,4,11,12], or by alternating between minimizing the reconstruction loss and clustering loss [13]. Autoencoders have also been used to encode self-expressiveness in subspace-based methods [14–16]. Lastly, there are probabilistic methods to deep clustering that employ variational autoencoders in various manners [17,18].

**Deep Embedded Clustering (DEC)** [3]. DEC is one of the first methods for clustering with deep neural networks, and a cornerstone method in deep clustering. It uses a multilayer perceptron (MLP), pre-trained as a stacked denoising autoencoder, and a clustering module based on soft-assignments to a set of centroids. DEC's loss function is constructed to force the distribution of soft cluster assignments closer to a target distribution, by means of the Kullback–Leibler (KL) divergence. The target distribution is constructed from the soft assignments, and designed to strengthen predictions and put more emphasis on high-confidence assignments. In Section 6.5, we investigate the performance of two variants of DEC, with and without the proposed UCOs.

Improved DEC (IDEC) [2] is a modification of the DEC algorithm where the reconstruction loss is kept during the optimization of the clustering objective. Guo et al. [2] argue that retaining the reconstruction loss leads to less corruption of the feature space, reducing in more representative features and better clustering performance.

### 2.2. Deep clustering without autoencoders

Several methods that do not use autoencoders in their architectures have also been developed in recent years. These include methods based on hierarchical clustering [19], subspace clustering [20–22], generative adversarial networks [23], and adversarial learning [24]. The recent success of self-supervised representation learning for images has also given rise to numerous methods for image clustering [25,26]. However, these methods depend heavily on image augmentations and are thus not directly applicable to other data types. Finally, there are approaches to deep clustering that do not rely on any auxiliary model components or losses [27,28]. In these models, the networks are trained end-to-end from random initializations by minimizing their respective clustering losses.

### 2.3. Objective function mismatch in autoencoder-based models

Metz et al. [29] demonstrate the presence of OFM in the context of unsupervised representation learning. They measure few-shot classification accuracy on representations obtained with a variational autoencoder, trained on the CIFAR-10 dataset. After an initial increase, they observe that the few-shot classification accuracy decreases in later stages of training the variational autoencoder. A more extensive study by Stuhr and Brauer [30] finds that OFM is present in several autoencoder-based models used for downstream image classification.

### 2.4. Objective function mismatch and feature drift in autoencoder-based deep clustering models

OFM was first observed in deep clustering models by Mrabah et al. [5,6]. In these works, they introduce *Feature Drift* (FD) as a measure to quantify OFM in deep clustering models. FD is defined as the cosine of the angle between the gradient of the clustering loss, and the gradient of the auxiliary loss, with respect to the model's weights. FD thus measures the negative "agreement" between the auxiliary loss and the clustering loss. Mrabah et al. [6] further show that the well-known Improved Deep Embedded Clustering (IDEC) [31] suffers from FD.

In order to address the FD issue in deep clustering, Mrabah et al. have developed the Adversarial Deep Embedded Clustering (ADEC) [6] and Dynamic Autoencoder (DynAE) [5] models.

ADEC [6] is a modification of IDEC [2], where an additional discriminator network, and adversarial loss is added to enhance the model's ability to reconstruct the input observations. Mrabah et al. argue that adding a discriminator and an adversarial loss allows the reconstruction loss to only act on the decoder weights during training, thereby reducing FD between the clustering and reconstruction tasks.

DynAE [5] is also a modification of IDEC [2], including a generator network and a critic, also aiming at improving the reconstruction capabilities of the model. DynAE reduces FD by gradually moving away from reconstructing individual observations, towards reconstruction the *closest centroid* of samples with high-confidence assignments.

We note that training both ADEC and DynAE requires data augmentation, adding another layer of complexity to the training procedure. Furthermore, it is not trivial to define suitable augmentation strategies for non-image data, such as vectors,[3] time-series, graphs, etc. ADEC and DynAE are found to reduce FD compared to other deep clustering methods. However, we find that there are several drawbacks related to using FD as a measure of OFM. These drawbacks are discussed further in Section 5.

In this work we propose the UCOs, which are alternative, non-augmentation-based auxiliary loss functions for deep clustering. The UCOs can be adopted to a broad class of deep clustering models without significant modifications to the base architecture, making them more versatile compared to ADEC and DynAE.

## 3. Unsupervised companion objectives

In deep clustering, the input observation $\mathbf{X}_i$, $i = 1, \dots, n$ is transformed by an encoder network $f$, producing a hidden representation $\boldsymbol{z}_i$. The cluster membership vector is then determined by a clustering module, $g$, as $\boldsymbol{\alpha}_i = g(\boldsymbol{z}_i)$.

Since $f$ is a neural network, we assume that it can be decomposed into consecutive computational *blocks*: $f = f^B \circ f^{B-1} \circ \cdots \circ f^1$, where $f^b$ denotes block $b$. A block can be, for instance, a single layer, or a collection of adjacent layers. We let $\mathbf{Y}_i^b = f^b(\mathbf{Y}_i^{(b-1)})$ be the output of block $b$ for observation $i$, where $\mathbf{Y}_i^0 = \mathbf{X}_i$. When we augment a deep clustering model with the UCOs, we attach companion objectives to the outputs of the blocks in the network, and minimize them alongside the main clustering objective. Fig. 1 illustrates a deep clustering model with the UCOs attached to the neural network.

When minimized, the UCOs should enforce clusters to be separable and compact at the output of their respective blocks. Since clustering losses often characterize clusters as separable and compact groups of points in space, the UCOs will align with these clustering objectives — thereby reducing the OFM between objectives. Inspired by Deep Divergence-based Clustering (DDC) [27], we encourage separability and compactness by maximizing an estimate of the Cauchy–Schwarz (CS) divergence between clusters [32]. For a set of $k$ clusters characterized by $k$ probability density functions $p_1, \dots, p_k$, the CS divergence between them at block $b$ is given by:

$$D_{cs}^b = -\log\left( \binom{k}{2}^{-1} \sum_{\substack{i=1 \\ j>i}}^{k} \frac{\int p_i(\boldsymbol{Y}^b) p_j(\boldsymbol{Y}^b) \mathrm{d}\boldsymbol{Y}^b}{\sqrt{\int p_i(\boldsymbol{Y}^b)^2 \mathrm{d}\boldsymbol{Y}^b \int p_j(\boldsymbol{Y}^b)^2 \mathrm{d}\boldsymbol{Y}^b}} \right) \tag{1}$$

where $\boldsymbol{Y}^b$ is the output of block $b$. Following common practice in deep clustering, we assume that the number of clusters, $k$, is known [2,3,27].[4]

$D_{cs}^b$ can be maximized by minimizing the argument of the logarithm. Using a Gaussian kernel density estimate gives the following UCO for block $b$ in the network:

$$\ell_b = \sum_{\substack{i=1 \\ j>i}}^{k} \frac{\binom{k}{2}^{-1} \sum_{l,m=1}^{n} \alpha_{li} \alpha_{mj} k_\sigma^{\mathrm{UCO}}(\boldsymbol{Y}_l^b, \boldsymbol{Y}_m^b)}{\sqrt{\sum_{l,m=1}^{n} \alpha_{li} \alpha_{mi} k_\sigma^{\mathrm{UCO}}(\boldsymbol{Y}_l^b, \boldsymbol{Y}_m^b) \sum_{l,m=1}^{n} \alpha_{lj} \alpha_{mj} k_\sigma^{\mathrm{UCO}}(\boldsymbol{Y}_l^b, \boldsymbol{Y}_m^b)}} \tag{2}$$

where $\alpha_{li}$ denotes element $i$ of the cluster membership vector $\boldsymbol{\alpha}_l$, and $k_\sigma^{\mathrm{UCO}}(\boldsymbol{Y}_l^b, \boldsymbol{Y}_m^b)$ is a Gaussian kernel evaluated at $(\boldsymbol{Y}_l^b, \boldsymbol{Y}_m^b)$. $\sigma$ denotes the kernel width. Collecting the UCOs for all blocks in the network gives the loss:

$$\mathcal{L}_{\mathrm{UCO}} = \sum_{b=1}^{B} w_b \ell_b = \lambda \sum_{b=1}^{B} \omega(b) \ell_b \tag{3}$$

where $w_b$ is the weight of the UCO attached to block $b$, and $B$ is the number of blocks. To avoid having to specify the weight for each block individually, we compute $w_b$ as $w_b = \lambda \cdot \omega(b)$. Here $\lambda$ is a hyperparameter that specifies the overall weight for the UCOs, and $\omega : \{1, \dots, B\} \to [0, 1]$ is a function that computes the relative weight of the UCO attached to block $b$.

The final loss used to train the model is then:

$$\mathcal{L}_{\mathrm{total}} = \mathcal{L}_{\mathrm{cluster}} + \mathcal{L}_{\mathrm{UCO}}. \tag{4}$$

**Connection to Deep Divergence-based Clustering** [27]. The UCOs are related to DDC [27] as they both minimize the CS divergence between clusters. In DDC however, the CS divergence is estimated with hidden representations that are computed within the clustering module. These hidden representations, $\boldsymbol{h}_i$ are obtained by transforming the encoder outputs, $\mathbf{Y}_i^B$ with a fully-connected layer. A second fully-connected layer is then applied to the hidden representations, producing the cluster membership vectors $\boldsymbol{\alpha}_i$.

DDC's clustering loss consists of three terms:

$$\mathcal{L}_{\mathrm{cluster}}^{\mathrm{DDC}} = \mathcal{L}_{\mathrm{DDC},1} + \mathcal{L}_{\mathrm{DDC},2} + \mathcal{L}_{\mathrm{DDC},3}. \tag{5}$$

[3] ADEC includes experiments on vector data, but do not describe how the augmentation is performed, nor is the code for ADEC publicly available.

[4] The case when $k$ is unknown, as in [26], is an interesting direction in deep clustering. However, it is beyond the scope of this paper.

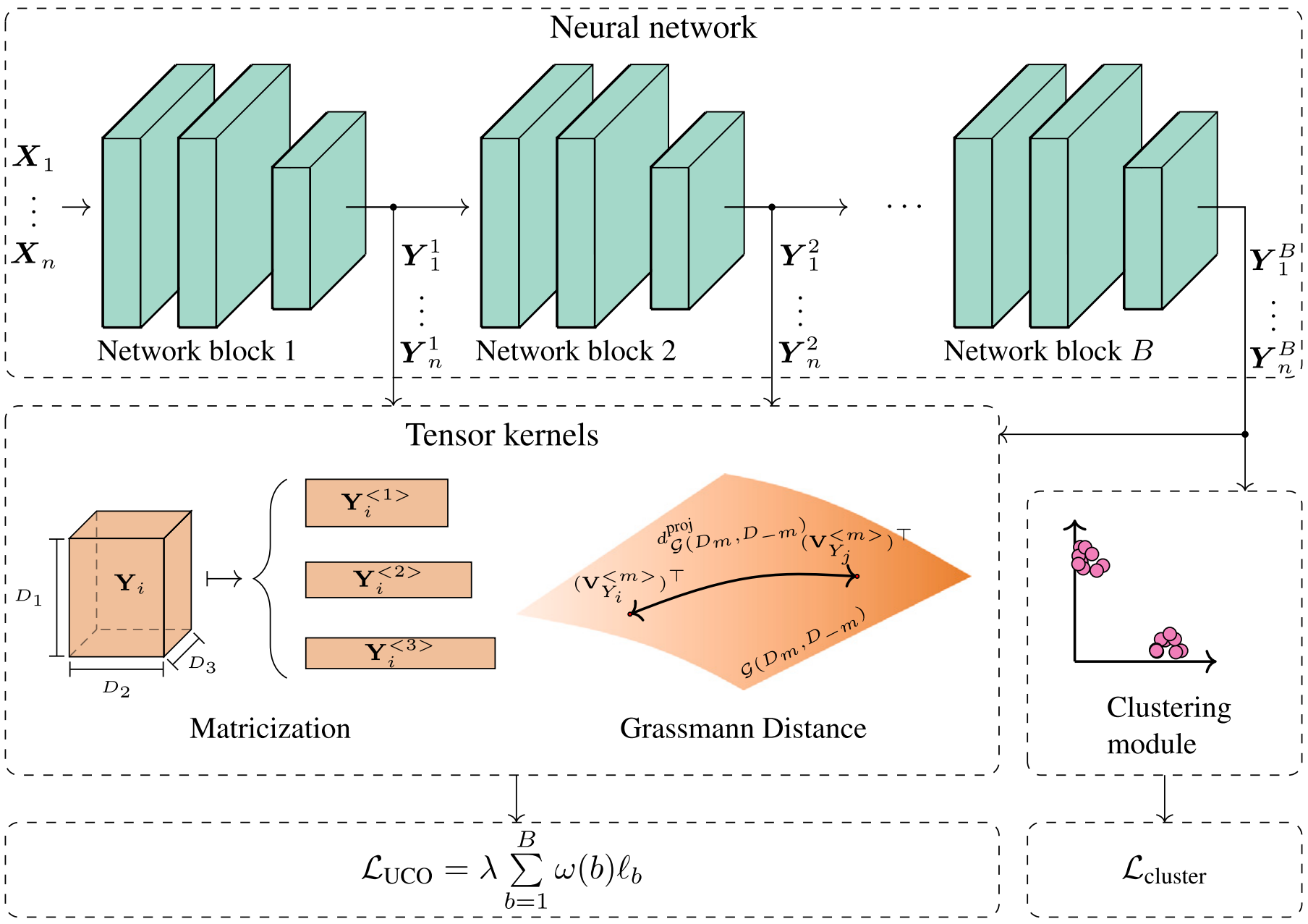

**Fig. 1.** Deep clustering model augmented with the proposed unsupervised companion objectives (UCOs). This model uses tensor kernels to estimate the CS-divergence between clusters at the block outputs. The estimate is then used in the computation of the UCOs.

The first term is similar to the UCOs, and minimizes the CS divergence between clusters in the space of hidden representations:

$$\mathcal{L}_{\text{DDC},1} = \sum_{\substack{i=1 \\ j>i}}^{k} \frac{\binom{k}{2}^{-1} \sum_{l,m=1}^{n} \alpha_{li}\alpha_{mj} k_\sigma(\boldsymbol{h}_l, \boldsymbol{h}_m)}{\sqrt{\sum_{l,m=1}^{n} \alpha_{li}\alpha_{mi} k_\sigma(\boldsymbol{h}_l, \boldsymbol{h}_m) \sum_{l,m=1}^{n} \alpha_{lj}\alpha_{mj} k_\sigma(\boldsymbol{h}_l, \boldsymbol{h}_m)}} \tag{6}$$

where $k$ is the number of clusters, $\alpha_{ij}$ is the soft cluster assignment for observation $i$ to cluster $j$, $\boldsymbol{h}_i$ is the hidden representation for the $i$th observation, and $k_\sigma$ is a Gaussian kernel with kernel width $\sigma$. As with the UCOs, minimizing this loss term results in clusters that are separable and compact in the hidden space.

The second term encourages the cluster membership vectors ($\boldsymbol{\alpha}_i$) for different observations to be orthogonal:

$$\mathcal{L}_{\text{DDC},2} = \sum_{i=1}^{n-1} \sum_{j=i+1}^{n} \boldsymbol{\alpha}_i^\top \boldsymbol{\alpha}_j. \tag{7}$$

The last term is also based on the CS divergence and pushes the cluster membership vectors closer to the corners of the standard simplex in $\mathbb{R}^k$:

$$\mathcal{L}_{\text{DDC},3} = \sum_{\substack{i=1 \\ j>i}}^{k} \frac{\binom{k}{2}^{-1} \sum_{l,m=1}^{n} \eta_{li}\eta_{mj} k_\sigma(\boldsymbol{h}_l, \boldsymbol{h}_m)}{\sqrt{\sum_{l,m=1}^{n} \eta_{li}\eta_{mi} k_\sigma(\boldsymbol{h}_l, \boldsymbol{h}_m) \sum_{l,m=1}^{n} \eta_{lj}\eta_{mj} k_\sigma(\boldsymbol{h}_l, \boldsymbol{h}_m)}} \tag{8}$$

where $\eta_{ij} = \exp(-\|\boldsymbol{\alpha}_i - \boldsymbol{e}_j\|^2)$, and $\boldsymbol{e}_j$ is a $k$-dimensional one-hot vector pointing to the $j$th corner of the simplex.

## 4. Tensor kernels

The UCOs in Eq. (2) depend on a kernel function ($k_\sigma^{\text{UCO}}$), which computes the similarity between block-outputs for different inputs. However, these outputs are not necessarily vectors. In CNNs for instance, the intermediate representations have two spatial dimensions, in addition to the feature dimension. The Gaussian kernel with Euclidean distance between vectors is thus not directly applicable to the block-outputs (without vectorizing them first). To address this issue, we consider the feature maps produced by a network block as rank-3 tensors. Building on Grassmannian learning [33] and tensor kernels [7], then allows us to specify a tensor kernel for the proposed UCOs.

### 4.1. *The naïve kernel*

For vectors we have the Gaussian kernel:

$$k_\sigma(\boldsymbol{x}, \boldsymbol{y}) = \exp\left(-\frac{\|\boldsymbol{x} - \boldsymbol{y}\|^2}{2\sigma^2}\right) = \prod_{i=1}^{D_1} \exp\left(-\frac{(x_i - y_i)^2}{2\sigma^2}\right) \tag{9}$$

where $x_i$ ($y_i$) refers to element $i$ in the vector $\boldsymbol{x}$ ($\boldsymbol{y}$). Generalizing this kernel to tensors $\mathbf{X}$ and $\mathbf{Y}$ of rank $r$ gives the naïve kernel [7]:

$$k_\sigma^{\text{naïve}}(\mathbf{X}, \mathbf{Y}) = \prod_{i_1=1}^{D_1} \cdots \prod_{i_r=1}^{D_r} \exp\left(-\frac{(X_{i_1 \cdots i_r} - Y_{i_1 \cdots i_r})^2}{2\sigma^2}\right) \tag{10}$$

where $X_{i_1 \cdots i_r}$ ($Y_{i_1 \cdots i_r}$) denotes the element of $\mathbf{X}$ ($\mathbf{Y}$) with indices $i_1, \ldots, i_r$, and $D_1, \ldots, D_r$ denote the number of elements along each dimension.

This kernel is equivalent to first vectorizing the representations, and then applying a Gaussian kernel using the Euclidean distance between the vectorized representations. However, the intermediate representations often lie on a low-dimensional manifolds where the Euclidean distance fails to be a good measure of dissimilarity. A potential reason for this is that the naïve kernel ignores the *structure* of the input tensors [7] – such as the spatial structure of the intermediate representations in CNNs. This means that the kernel is invariant to a fixed permutation rule $P$:

$$k_\sigma^{\text{naïve}}(\mathbf{X}, \mathbf{Y}) = k_\sigma^{\text{naïve}}(P(\mathbf{X}), P(\mathbf{Y})). \tag{11}$$

This effect can be especially destructive for images, for instance, since images can be transformed beyond recognition by permuting the spatial indices.

### 4.2. *Structure-exploiting tensor kernels*

The shortcomings of the naïve kernel outlined above calls for a kernel that takes the tensors' structure into account. The kernels are therefore formulated using the *matricizations* of input tensors. The matricization of a tensor $\mathbf{X}$ along dimension $m$ is obtained by vectorizing along all dimensions of $\mathbf{X}$, except dimension $m$. This results in a matrix with shape $(D_m, D_{-m})$ where $D_{-m} = \frac{1}{D_m} \prod_{l=1}^{r} D_l$ (see Fig. 2).

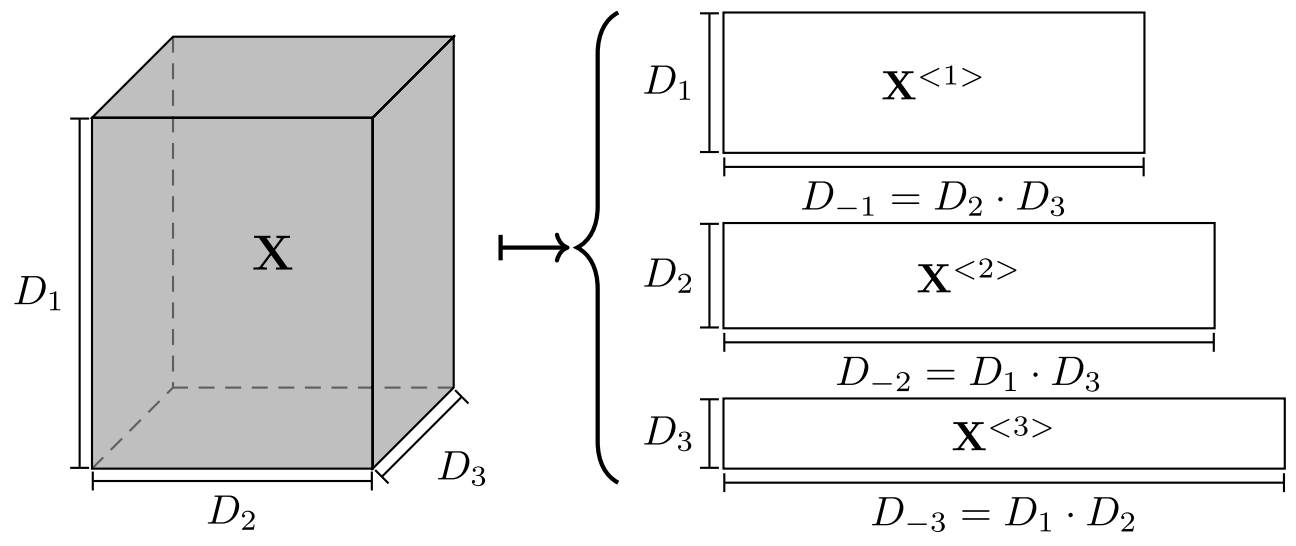

**Fig. 2.** Matricization of a rank-3 tensor **X**.

The matricizations along the tensor dimensions can then be used to form the tensor kernel:

$$k_\sigma^{\text{tensor}}(\mathbf{X}, \mathbf{Y}) = \prod_{m=1}^{r} k_\sigma^m(\mathbf{X}^{\langle m\rangle}, \mathbf{Y}^{\langle m\rangle}) \tag{12}$$

$$= \prod_{m=1}^{r} \exp\left(-\frac{d_{\mathcal{G}(D_m, D_{-m})}(\mathbf{X}^{\langle m\rangle}, \mathbf{Y}^{\langle m\rangle})^2}{2\sigma^2}\right) \tag{13}$$

where $\mathbf{X}^{\langle m\rangle}$ ($\mathbf{Y}^{\langle m\rangle}$) is the matricization of $\mathbf{X}$ ($\mathbf{Y}$) along dimension $m$. In order to specify a suitable distance function between the matricizations, we consider the span of their rows as points on the Grassmann manifold, $\mathcal{G}(D_m, D_{-m})$, which consists of all $D_m$ dimensional linear subspaces of $\mathbb{R}^{D_{-m}}$. This is a central concept in *Grassmannian learning* [34] – a field which has shown promising results, but is still in its early stages. Specifically, we exploit the one-to-one correspondence between linear subspaces and their orthogonal projection operators to define our distance function on the Grassmannian manifold [7]. Fig. 3 illustrates the distance function on the Grassmann manifold. The orthogonal projection operator is given by

$$\mathbf{\Pi}_X^{\langle m\rangle} = \mathbf{V}_X^{\langle m\rangle}(\mathbf{V}_X^{\langle m\rangle})^\top \tag{14}$$

where $\mathbf{V}_X^{\langle m\rangle}$ is obtained through the singular value decomposition (SVD):

$$\mathbf{X}^{\langle m\rangle} = \mathbf{U}_X^{\langle m\rangle}\mathbf{\Sigma}_X^{\langle m\rangle}(\mathbf{V}_X^{\langle m\rangle})^\top. \tag{15}$$

Considering the Frobenius norm between projection operators then gives the distance function:

$$d^{\text{proj}}_{\mathcal{G}(D_m, D_{-m})}(\mathbf{X}^{\langle m\rangle}, \mathbf{Y}^{\langle m\rangle})$$

$$= \|\mathbf{\Pi}_X^{\langle m\rangle} - \mathbf{\Pi}_Y^{\langle m\rangle}\|_F \tag{16}$$

$$= \sqrt{2(D_m - \text{Tr}((\mathbf{V}_Y^{\langle m\rangle})^\top \mathbf{V}_X^{\langle m\rangle}(\mathbf{V}_X^{\langle m\rangle})^\top \mathbf{V}_Y^{\langle m\rangle}))}. \tag{17}$$

where $\|\cdot\|_F$ denotes the Frobenius norm. Inserting $d^{\text{proj}}_{\mathcal{G}(D_m, D_{-m})}(\mathbf{X}^{\langle m\rangle}, \mathbf{Y}^{\langle m\rangle})$ into (13) allows us to compute a structure-exploiting tensor kernel $k_\sigma^{\text{tensor}}(\mathbf{X}, \mathbf{Y})$, which can then be included in the proposed UCOs by setting $k_\sigma^{\text{UCO}} = k_\sigma^{\text{tensor}}$ in (2).

We note that the projection distance gives rise to a positive semi-definite kernel. This allows the UCOs to be interpreted as a cosine distance between the mean of intermediate representations from different clusters, in the Reproducing Kernel Hilbert Space induced by the tensor kernel in Eq. (13). However, a thorough analysis of this property is left to future work.

## 5. Measuring objective function mismatch

### 5.1. Drawbacks of using feature drift as a measure of objective function mismatch

Feature Drift (FD) is defined as the cosine of the angle between the gradient of the main loss and the auxiliary loss [5,6]

$$\Delta_{FD} = \cos(\nabla_\theta \mathcal{L}_{\text{cluster}}, \nabla_\theta \mathcal{L}_{\text{aux}}) \tag{18}$$

where $\nabla_\theta$ denotes the gradient operator w.r.t. the model parameters $\theta$ (see Fig. 4 for illustration of low FD vs. high FD). $\Delta_{FD}$ measures the *agreement* between the losses $\mathcal{L}_{\text{cluster}}$ and $\mathcal{L}_{\text{aux}}$, meaning that *higher values of $\Delta_{FD}$ indicate less OFM.* Hence, when using $\Delta_{FD}$ to measure OFM, we should seek a solution where $\Delta_{FD}$ is maximized, in order to minimize OFM. Based on these observations, we identify the following drawbacks of using FD to measure OFM:

1. From Eq. (18) we observe that setting $\mathcal{L}_{\text{aux}} = c\mathcal{L}_{\text{cluster}}$, for some positive real constant $c$, we obtain $\Delta_{FD} = 1$. This means that according to the FD measure, OFM can be trivially eliminated by making the auxiliary loss proportional to the clustering loss. Although it is natural that the mismatch between objective functions can be eliminated by making the objectives equal, this is not a particularly helpful result for designing new clustering methods.
   This result also underlines another key property about OFM in deep clustering, namely that there might exist a positive amount of OFM that is beneficial for the clustering task – *i.e.* there is exists a "right amount" of OFM that can be different from 0. In classification models for instance, $L_2$ regularization is found to improve classification performance, even though the regularizer most likely has some degree of mismatch between it and the main classification loss.
2. $\Delta_{FD}$ does not take the gradient norms into account. In the event that $\|\nabla_\theta \mathcal{L}_{\text{cluster}}\| \gg \|\nabla_\theta \mathcal{L}_{\text{aux}}\|$, $\Delta_{FD}$ can be arbitrary high or low, even though the influence of the auxiliary loss is negligible, due to the low gradient norm.
3. $\Delta_{FD}$ is essentially an inner product between unit vectors in a space with dimension equal to the number of parameters in the model. However, the distribution of inner products between uniform random unit vectors will have vanishing mean and variance as the dimension goes to $\infty$ [35]. The variance of $\Delta_{FD}$ can therefore vanish for large models, making the measure less informative.

### 5.2. R-OFM: A regression-based measure of objective function mismatch

In order to address the drawbacks of FD described above, we develop a new, regression-based measure of OFM. Crucially, our new measure, R-OFM (i) Does not rely on gradient computations, and consequently does not discard potentially relevant information encoded in gradient norms. (ii) Directly measures relationship between auxiliary loss and clustering performance. (iii) Cannot be trivially minimized by setting the auxiliary loss equal to the clustering loss.

Suppose $a_t$ is the value of the auxiliary loss at training iteration $t$, and $p_t$ is a measure of clustering performance[5] at the same iteration. Given pairs $(a_t, p_t)$ measured during training, we first formulate the following regression model

$$p_t = \varepsilon_t + \sum_{i=0}^{\gamma} \beta_i a_t^i, \quad t = 1, \dots, T \tag{19}$$

where $\varepsilon_t \sim \mathcal{N}(0, s)$ is a Gaussian error term, and $\gamma$ is a hyperparameter determining the degree of the polynomial regressor.[6] Estimates of the regression parameters, $\hat{\beta}_0, \dots, \hat{\beta}_\gamma$, can then be obtained with *e.g.* ordinary least squares. We then compute two values from the fitted regression model:

1. The average slope of the regression line, evaluated at the observed loss values $a_1, \dots, a_T$

$$\text{R-OFM}_{\text{slope}} = \frac{1}{T}\sum_{t=1}^{T}\sum_{i=1}^{\gamma} i \cdot \hat{\beta}_i \cdot a_t^{i-1} \tag{20}$$

[5] *E.g.* clustering accuracy.
[6] We set $\gamma = 2$ in our experiments, as it allows the regression model to be sufficiently flexible, while keeping the number of parameters low.

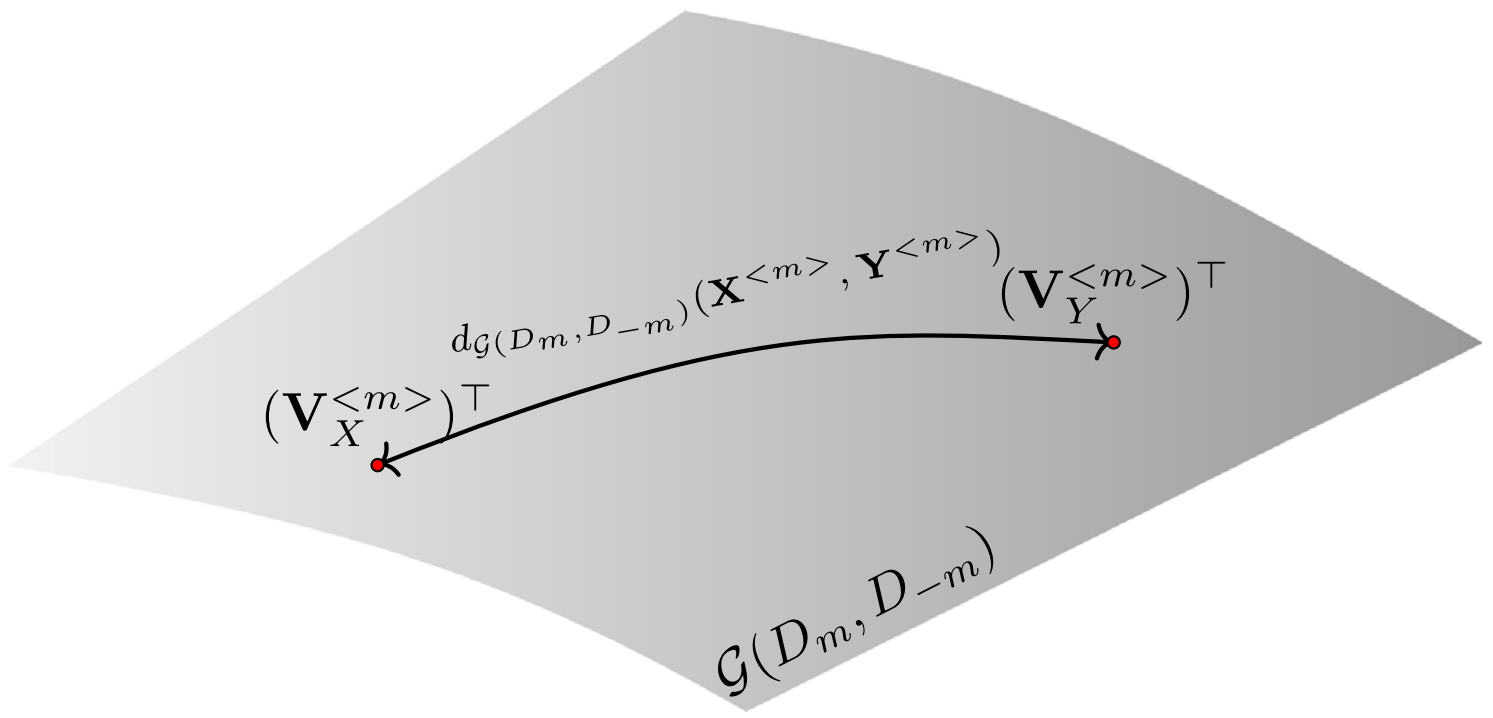

**Fig. 3.** The distance $d_{\mathcal{G}(D_m,D_{-m})}(\mathbf{X}^{\langle m\rangle},\mathbf{Y}^{\langle m\rangle})$ between two matricizations $\mathbf{X}^{\langle m\rangle}$ and $\mathbf{Y}^{\langle m\rangle}$ on the Grassmann manifold $\mathcal{G}(D_m,D_{-m})$. Since the Grassmann manifold consists of linear subspaces, the matricizations are represented by the respective orthonormal representations $(\mathbf{V}_X^{\langle m\rangle})^\top$ and $(\mathbf{V}_Y^{\langle m\rangle})^\top$.

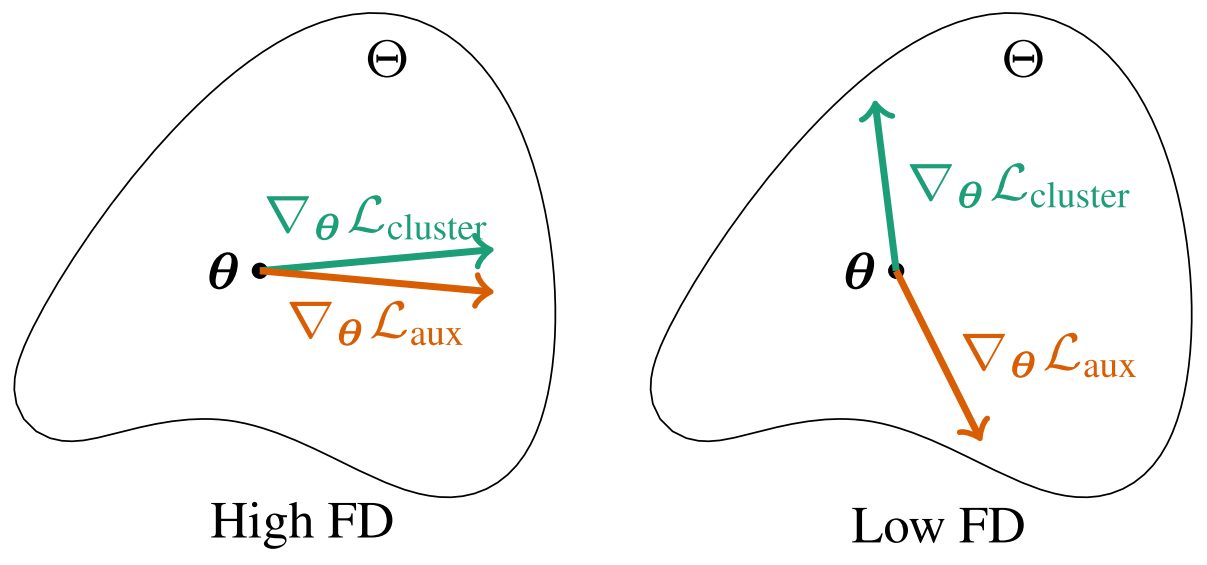

**Fig. 4.** Illustration of FD for different loss gradients. FD is defined as the cosine of the angle between the gradient of the clustering loss and the auxiliary loss wrt. the encoder parameters $\theta \in \Theta$.

2. The model's $R^2$ value, indicating the goodness-of-fit of the model

$$\text{R-OFM}_{R^2} = \frac{\sum_{t=1}^{T}(p_t - \hat{p}_t)^2}{\sum_{t=1}^{T}(p_t - \bar{p})^2} \tag{21}$$

where $\hat{p}_t = \sum_{i=0}^{\gamma} \hat{\beta}_i a_t^i$ is the predicted value of the performance at iteration $t$, and $\bar{p} = \frac{1}{T}\sum_{t=1}^{T} p_t$ is the mean performance.

These two metrics together makes it possible to accurately estimate the amount of OFM present while training a model. $\text{R-OFM}_{\text{slope}}$ measures how much the clustering performance changes as a function of the auxiliary loss. A *negative* value of $\text{R-OFM}_{\text{slope}}$ indicates that the performance increases as the loss decreases, indicating that minimizing the auxiliary loss leads to improved clustering performance. $\text{R-OFM}_{R^2}$ complements $\text{R-OFM}_{\text{slope}}$ by indicating how strongly the auxiliary loss and clustering performance are correlated. A *high* value of $\text{R-OFM}_{R^2}$ indicates that there is a strong relationship between auxiliary loss and clustering performance.

## 6. Experiments

In this section we report the effects of augmenting two different deep clustering models with the proposed UCOs — both in terms of clustering performance, and the OFM between the main clustering objective and the auxiliary objective. Further, we investigate the effect of varying the weighting hyperparameters, $\omega$ and $\lambda$, in the UCOs.

### *6.1. Setup*

**Models.** Our experiments are based on DDC [27], as it is an excellent general-purpose deep clustering method, which has demonstrated state-of-the-art clustering performance on images [27], time series [36], and several types of multi-view data [37]. In Section 6.5 we also show some results with DEC-based models.

We augment DDC with UCOs based on tensor kernels, and refer to the resulting model as DDC-UCO$_T$. To evaluate the effects of the structure-exploiting tensor kernels, we also include a model with UCOs based on vectorization and naïve kernels, referred to as DDC-UCO$_N$.

In addition, we train DDC with a decoder and a reconstruction loss (we refer to this model as DDC-AE). This allows us to accurately compare DDC-UCO$_{N/T}$ to an analogous autoencoder-based model.

Finally, we include results with DynAE [5], to compare the proposed methodology to previous work on OFM reduction for deep clustering, and with Pseudo-supervised Deep Subspace Clustering (PSSC) [16] to compare to a recent state-of-the-art method for deep clustering.

We train the models with randomly initialized parameters, using their respective loss functions. DynAE and PSSC were trained using the code and hyperparameters provided by Mrabah et al. [6][7] and Lv et al. [16][8], respectively.

For the DDC-based models, the batch size is set to 120, and we use the Adam optimizer with default parameters. Following [27], we let the $\sigma$ hyperparameter be 15% of the median distance between observations in a batch. This rule of thumb is used for both the UCOs and the clustering module.

Due to the difficulty of hyperparameter validation in unsupervised learning, we compute the UCO weights with $\omega(b) = 1$ (constant) and $\lambda = 0.01$ for all datasets. This choice is further studied in Section 6.4.

**Datasets.** We test the models on 6 different benchmark datasets:

1. MNIST [38]: Grayscale images of handwritten digits.
2. Fashion-MNIST [39]: Grayscale images of clothing items.
3. COIL-100 [40]: RGB Images of 100 common objects depicted from 72 different angles.
4. USPS: Grayscale images of handwritten digits.
5. Reuters [41]: TF-IDF features from news stories.[9]
6. $10 \times 73$k [42]: Feature vectors for RNA-transcripts from different cell types.

More details on the datasets can be found in Table 1

**Evaluation.** We train each model 20 times per dataset, and report the performance of the run which resulted in the lowest value of the total loss function (Eq. (4)). This is the same evaluation procedure as in DDC [27].

To measure the clustering performance of the different algorithms, we use the unsupervised clustering accuracy (ACC) and the normalized mutual information (NMI). Both ACC and NMI are bounded in $[0, 1]$, and higher values indicate better clusterings, with respect to the ground-truth labels.

[7] https://github.com/nairouz/DynAE.

[8] https://github.com/sckangz/SelfsupervisedSC.

[9] As in [27], we select 54 000 samples from each class to balance the dataset.

**Table 1**
Details of the benchmark datasets.

| Name | Type | Dimension | Samples | Classes |
|---|---|---|---|---|
| MNIST [38] | Images | $28 \times 28 \times 1$ | 60 000 | 10 |
| Fashion-MNIST [39] | Images | $28 \times 28 \times 1$ | 60 000 | 10 |
| COIL-100 [40] | Images | $128 \times 128 \times 3$ | 7200 | 100 |
| USPS | Images | $16 \times 16 \times 1$ | 9298 | 10 |
| Reuters [41] | Vectors | 2000 | 216 000 | 4 |
| 10 × 73k [42] | Vectors | 720 | 73 233 | 8 |

**Table 2**
Clustering results for DDC-based models on the benchmark datasets. Standard deviations obtained by bootstrapping are shown in parentheses. The best result for each base model is highlighted in **bold**, and results that are within one standard deviation of the best are <u>underlined</u>.

| | MNIST | | F-MNIST | |
|---|---|---|---|---|
| | ACC | NMI | ACC | NMI |
| DDC | 88.2 (2.1) | 87.7 (2.3) | 48.6 (8.1) | 44.3 (6.0) |
| DDC-AE | 87.9 (1.3) | 86.2 (0.7) | 60.3 (4.7) | 55.1 (3.8) |
| DynAE | 83.0 (2.3) | 84.3 (0.9) | 49.0 (2.9) | 56.6 (0.8) |
| PSSC | 72.8 (1.2) | 74.7 (0.6) | 60.3 (2.4) | **63.5** (1.0) |
| DDC-UCO$_N$ | 85.5 (4.7) | 82.4 (3.3) | **65.5** (5.0) | 59.1 (3.6) |
| DDC-UCO$_T$ | **96.0** (6.4) | **91.7** (5.2) | <u>65.3</u> (4.1) | 59.2 (3.4) |
| | COIL-100 | | USPS | |
| | ACC | NMI | ACC | NMI |
| DDC | 58.0 (1.1) | 82.6 (0.2) | 67.9 (2.8) | 70.1 (1.6) |
| DDC-AE | 61.2 (1.3) | **84.4** (0.2) | 72.1 (5.4) | 71.0 (4.9) |
| DynAE | – | – | **84.9** (0.9) | **84.9** (0.3) |
| PSSC | 52.8 (1.1) | 81.5 (0.3) | 73.3 (2.6) | 81.6 (1.6) |
| DDC-UCO$_N$ | 58.5 (1.4) | 80.8 (0.3) | 75.5 (4.8) | 76.6 (3.3) |
| DDC-UCO$_T$ | **62.9** (1.1) | 83.8 (0.5) | 75.6 (3.1) | 77.4 (2.5) |
| | Reuters | | 10 × 73k | |
| | ACC | NMI | ACC | NMI |
| DDC | 50.9 (3.6) | 24.2 (5.8) | 62.3 (1.0) | 55.2 (2.3) |
| DDC-AE | 37.2 (5.3) | 12.1 (6.1) | <u>78.6</u> (3.7) | <u>74.4</u> (3.1) |
| DDC-UCO$_N$ | 55.5 (4.8) | 30.4 (4.2) | 73.1 (3.2) | 70.2 (1.9) |
| DDC-UCO$_T$ | **61.2** (5.3) | **35.6** (4.8) | **81.1** (1.4) | **76.3** (0.9) |

### 6.2. Clustering results

The clustering results in Table 2 show that adding the UCOs to DDC improves the overall performance of the model. Furthermore, DDC-UCO$_T$ consistently outperforms DDC-UCO$_N$, indicating that the structure-exploiting tensor kernels are suitable for quantifying the similarities between the intermediate representations of the network.

Interestingly, we also observe that adding a decoder and reconstruction loss to DDC only improves ACC and NMI on COIL-100, as well as the NMI on USPS. On MNIST and F-MNIST, the performance of DDC-AE is significantly worse than the original DDC, indicating that the addition of an autoencoder can be harmful to the performance of DDC.

The results show that PSSC [16] is outperformed by DDC-UCO$_{N/T}$ in terms of accuracy on all image datasets. In terms of NMI, PSSC performs better than DDC-UCO$_{N/T}$ on USPS and F-MNIST, but is outperformed on the other image datasets. For DynAE [5], we observe that DDC-UCO$_{N/T}$ outperforms DynAE on MNIST and F-MNIST, whereas DynAE performs best on USPS. USPS has much fewer samples than MNIST and F-MNIST (9298 vs. 60 000) – likely causing the data augmentation to be more important for USPS than of MNIST/F-MNIST. Note that we were only able to run DynAE on USPS, MNIST and F-MNIST, and PSSC on USPS, MNIST, F-MNIST and COIL-100, since these were the datasets supported by the public implementations. PSSC uses a convolutional autoencoder and expects image data as input. For DynAE, it is not trivial to formulate alternative data augmentation strategy for the vector datasets Reuters and 10 × 73k.

**Table 3**
OFM metrics (R-OFM$_{\text{slope}}$ and R-OFM$_{R^2}$) for different datasets and models. For R-OFM$_{\text{slope}}$ lower values are better, and for R-OFM$_{R^2}$ higher values are better.

| | MNIST | | F-MNIST | |
|---|---|---|---|---|
| | R-OFM$_{\text{slope}}$ | R-OFM$_{R^2}$ | R-OFM$_{\text{slope}}$ | R-OFM$_{R^2}$ |
| DDC-AE | −0.092 | 0.370 | 0.058 | 0.067 |
| DynAE | 0.000 | 0.716 | −0.003 | 0.431 |
| DDC-UCO$_N$ | −0.089 | 0.575 | −0.057 | 0.073 |
| DDC-UCO$_T$ | −0.816 | 0.573 | −0.186 | 0.712 |
| | COIL-100 | | USPS | |
| | R-OFM$_{\text{slope}}$ | R-OFM$_{R^2}$ | R-OFM$_{\text{slope}}$ | R-OFM$_{R^2}$ |
| DDC-AE | −0.029 | 0.145 | 0.043 | 0.053 |
| DynAE | – | – | −0.114 | 0.465 |
| DDC-UCO$_N$ | −0.024 | 0.963 | −0.157 | 0.556 |
| DDC-UCO$_T$ | −0.004 | 0.761 | −0.067 | 0.760 |
| | Reuters | | 10 × 73k | |
| | R-OFM$_{\text{slope}}$ | R-OFM$_{R^2}$ | R-OFM$_{\text{slope}}$ | R-OFM$_{R^2}$ |
| DDC-AE | 0.797 | 0.603 | −0.029 | 0.026 |
| DDC-UCO$_N$ | 0.067 | 0.087 | −0.200 | 0.600 |
| DDC-UCO$_T$ | −0.082 | 0.232 | −0.000 | 0.114 |

Fig. 5 shows intermediate MNIST representations after the first block, projected to 2 dimensions with $t$-SNE [43]. From these plots we make the following observations:

- The clusters of 4, 5 and 7 are all more compact for the representations produced by DDC-UCO$_T$, compared to the others.
- Similar looking digits, such as 4 and 9, as well as 3, 5, and 8, are better separated in the DDC-UCO$_T$ representations, compared to the other models.

These observations illustrate that the UCOs indeed encourage compact and separable clusters in the space of intermediate representations. This results in better separability for the challenging cases where different digits look similar to each other.

### 6.3. Objective function mismatch

Table 3 shows values of R-OFM$_{\text{slope}}$ and R-OFM$_{R^2}$ for the different models and datasets. Overall we observe that DDC-UCO$_{N/T}$ has lower R-OFM$_{\text{slope}}$ and higher R-OFM$_{R^2}$, indicating less OFM and better correspondence between improved performance and decreasing auxiliary loss.

Furthermore, comparing Tables 2 and 3, we observe a correspondence between low OFM and high performance *across models*, meaning that models with less OFM tend to perform better than models with high OFM. However, the results do not show a complete one-to-one correspondence between OFM and accuracy. On USPS, DynAE has marginally more OFM compared to DDC-UCO$_N$, but performs better.

Lastly, we observe that DynAE has the highest value of R-OFM$_{R^2}$, but also has R-OFM$_{\text{slope}}$ $= 0$, implying that minimizing the auxiliary objective has no impact on clustering performance. This underlines that both R-OFM$_{\text{slope}}$ and R-OFM$_{R^2}$ has to be taken into account when considering the OFM exhibited by a model.

Note that PSSC [16] could not be included in the OFM analysis since clustering is done with spectral clustering *after* the model has been trained.

### 6.4. UCO weighting strategy

To validate our chosen UCO weighting strategy (choice of $\omega$ and $\lambda$), we train DDC-UCO$_{N/T}$ with $\lambda \in \{0.001, 0.01, 0.1\}$, and the following three $\omega$ functions: (i) $\omega(b) = 1$ (Constant); (ii) $\omega(b) = 10^{B-b}$ (Exponential); and (iii) $\omega(b) = b/B$ (Linear).

The resulting clustering accuracies for the different configurations are listed in Table 4. These results show that DDC-UCO$_{N/T}$ are not sensitive to the choice of $\omega$ and $\lambda$. This is an important property as labeled validation data is generally not available in unsupervised learning.

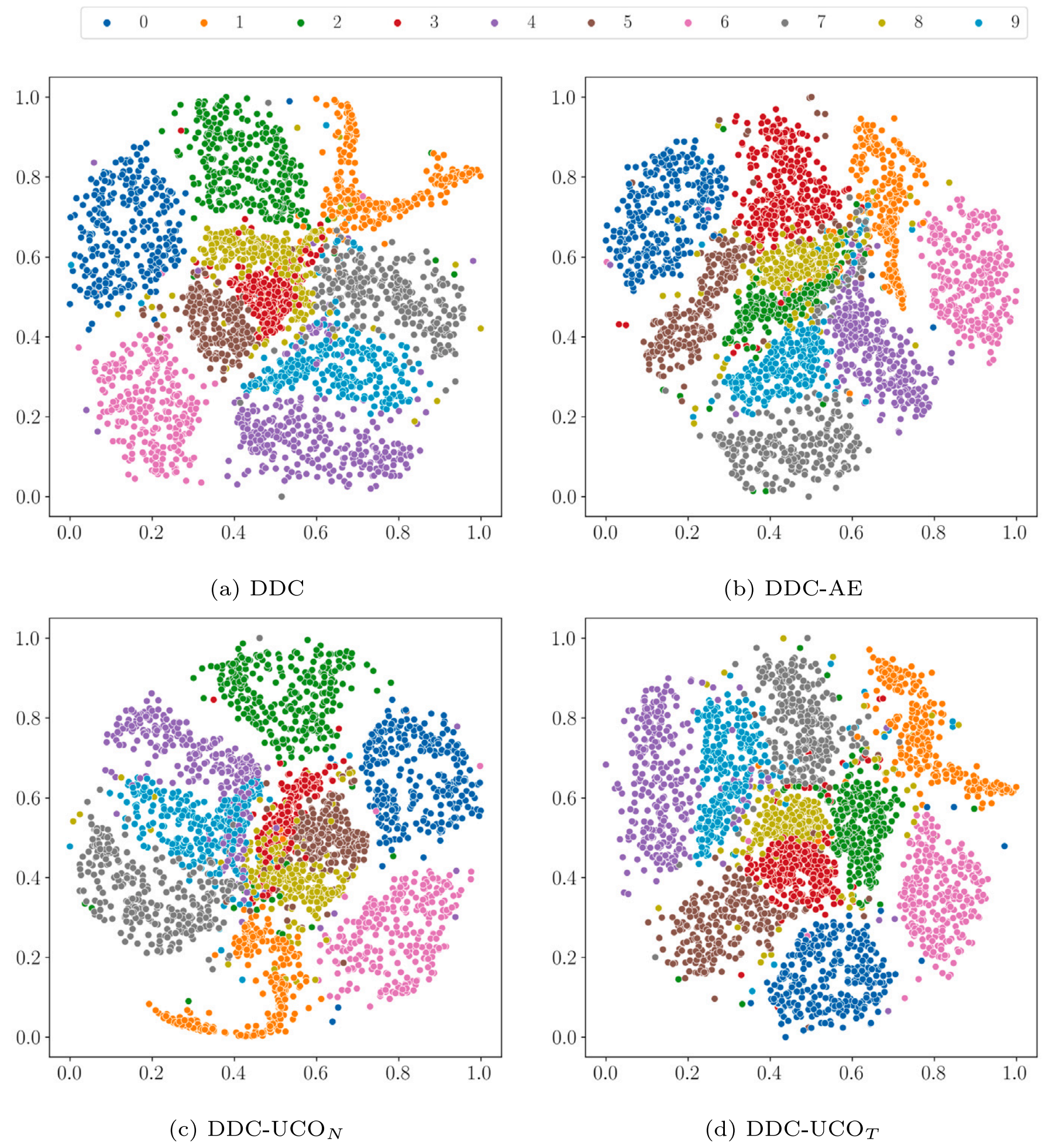

**Fig. 5.** $t$-SNE [43] plot of intermediate MNIST representations from the first network block $(\mathbf{Y}_1^1, \dots, \mathbf{Y}_n^1)$.

**Table 4**
Clustering results (ACC) for DDC-UCO$_{N/T}$ with different UCO weighting methods.

| Dataset | $\omega(\cdot) \downarrow$ $\lambda \rightarrow$ | DDC-UCO$_N$ | | | DDC-UCO$_T$ | | |
|---|---|---|---|---|---|---|---|
| | | 0.001 | 0.010 | 0.100 | 0.001 | 0.010 | 0.100 |
| MNIST | Constant | 85.2 | 85.5 | 87.4 | 86.6 | **96.0** | 87.2 |
| | Exponential | **88.2** | 87.1 | 87.1 | 87.3 | 85.7 | 87.7 |
| | Linear | 76.0 | 81.3 | 82.3 | 87.8 | 87.6 | 88.6 |
| COIL-100 | Constant | 60.5 | 58.5 | 63.1 | 59.6 | 62.9 | 61.7 |
| | Exponential | 59.0 | 60.4 | 60.8 | **65.9** | 63.0 | 62.7 |
| | Linear | 61.9 | **63.7** | 60.5 | 60.1 | 61.4 | 58.5 |
| F-MNIST | Constant | 61.8 | **65.5** | 49.9 | 59.5 | **65.3** | 51.0 |
| | Exponential | 52.8 | 51.7 | 60.0 | 59.1 | 59.1 | 60.1 |
| | Linear | 55.4 | 59.1 | 46.1 | 59.8 | 60.3 | 50.0 |
| USPS | Constant | 65.4 | 75.5 | 66.8 | 69.6 | **75.6** | 72.9 |
| | Exponential | 69.4 | 58.3 | 66.3 | 72.6 | 66.8 | 73.3 |
| | Linear | 66.5 | 69.3 | **76.1** | 74.4 | 73.5 | 69.6 |
| Reuters | Constant | 45.8 | 54.4 | 66.3 | 58.4 | 55.1 | 62.8 |
| | Exponential | 60.2 | 66.8 | **75.6** | 50.0 | 57.2 | 62.7 |
| | Linear | 48.1 | 63.1 | 50.6 | 60.3 | **71.7** | 58.3 |
| 10 × 73k | Constant | 82.4 | 73.1 | 87.9 | 78.4 | 81.1 | **86.9** |
| | Exponential | 81.2 | 84.6 | **89.2** | 78.8 | 70.4 | 86.2 |
| | Linear | 82.6 | 78.8 | 78.0 | 81.7 | 82.1 | 85.1 |

We note that, for some configurations on F-MNIST and Reuters, the performance degrades significantly when $\lambda = 0.1$ or $\lambda 0.001$, indicating that the influence of the UCOs is too strong or too weak in these cases.

### 6.5. Experiments with other base models

In order to evaluate the effect of the UCOs on other deep clustering models, we implement two variants of the well-known DEC model [3]. The first is the original DEC model with an MLP encoder. The second model, DEC-Conv, uses the same clustering module as DEC, but has a CNN encoder instead. We also train these variants with a decoder and reconstruction loss during fine-tuning (as is done in Improved DEC [2] and Deep Convolutional Embedded Clustering (DCEC) [31], respectively). Finally, we create two new models by augmenting the two variants with our UCOs. Since CNNs are designed for images and MLPs are designed for vectors, we evaluate DEC-Conv on the image datasets (MNIST, F-MNIST, COIL-100, USPS), and DEC on the vector datasets (Reuters, 10 × 73k).

The results in Table 5 show that the DEC-based models also benefit from the UCOs. This indicates that the UCOs work well with different clustering modules — not only DDC. However, the improvement in clustering performance is not as large for DEC and DEC-Conv, as for DDC. This discrepancy is likely due to the difference between DEC's clustering objective, and the UCOs. In DDC, the clustering objective is partially based on the CS divergence between clusters. This is the same divergence measure that we use in the UCOs. DEC's clustering objective however, is based on a Kullback–Leibler divergence between densities of cluster assignments. It is therefore more dissimilar to the UCOs, compared to the clustering objective in DDC.

**Table 5**
Clustering results with the models based on DEC-Conv (image datasets) and DEC (vector datasets). Standard deviations obtained by bootstrapping are shown in parentheses. The best result for each base model is highlighted in **bold**, and results that are within one standard deviation of the best are underlined.

| | MNIST | | F-MNIST | |
|---|---|---|---|---|
| | ACC | NMI | ACC | NMI |
| DEC-Conv | 78.3 (5.3) | 73.6 (2.8) | 54.7 (1.4) | 60.1 (0.9) |
| DCEC | 77.8 (1.3) | 72.6 (1.2) | 54.3 (2.7) | 60.0 (0.8) |
| DEC-Conv-UCO$_N$ | 80.0 (1.1) | 77.0 (0.5) | 57.0 (2.9) | 60.8 (1.0) |
| DEC-Conv-UCO$_T$ | **81.2** (0.8) | **79.1** (0.6) | **59.8** (3.5) | **62.0** (1.3) |
| | COIL-100 | | USPS | |
| | ACC | NMI | ACC | NMI |
| DEC-Conv | **50.3** (3.7) | **74.5** (2.2) | 73.6 (0.3) | 73.1 (0.2) |
| DCEC | 47.6 (2.7) | 71.7 (2.0) | 75.3 (0.5) | 75.5 (0.6) |
| DEC-Conv-UCO$_N$ | 42.9 (2.5) | 70.0 (1.8) | 76.9 (0.7) | 78.1 (1.1) |
| DEC-Conv-UCO$_T$ | 38.9 (0.7) | 68.0 (1.0) | **78.0** (0.6) | **80.7** (0.9) |
| | Reuters | | 10 × 73k | |
| | ACC | NMI | ACC | NMI |
| DEC | 68.6 (1.1) | **41.7** (1.7) | 74.9 (0.7) | 74.6 (0.8) |
| IDEC | 66.1 (1.1) | 35.9 (2.5) | 73.0 (0.6) | 73.2 (0.7) |
| DEC-UCO$_N$ | 64.1 (2.9) | 35.4 (3.5) | 71.4 (0.4) | 69.5 (0.6) |
| DEC-UCO$_T$ | **69.3** (0.8) | 40.0 (1.1) | **75.8** (0.7) | **76.2** (1.1) |

## 7. Limitations and future work

The over-arching challenge with OFM in deep clustering is to design models with "the right amount" of OFM — correctly balancing the main and auxiliary losses to achieve the best performance. To this end, we have developed a new measure of OFM that aligns well with clustering performance. In our experiments we use the clustering accuracy to compute R-OFM$_{\text{slope}}$ and R-OFM$_{R^2}$, which can be problematic in real-world applications where ground-truth test data is unavailable. Thus, a possible direction for future work is to incorporate unsupervised performance metric, such as internal cluster validity indices, in order to make R-OFM$_{\text{slope}}$ and R-OFM$_{R^2}$ label-independent.

In this work we have also proposed a set of new auxiliary objectives to reduce OFM and improve performance. Although the tensor kernels make the UCOs more versatile and better performing, they introduce an additional computational overhead. This overhead mainly comes from the SVD computations (Eq. (15)), and could possibly be reduced by considering other distance functions on the grassmannian manifold.

## 8. Conclusion

This paper sheds new light on the problem of objective function mismatch (OFM) in deep clustering. The results of our experiments show that coupling a deep clustering model with an autoencoder can cause a significant amount of mismatch between the clustering objective and the reconstruction objective — possibly leading to reduced clustering performance.

To address the issue of OFM in deep clustering, we introduced the unsupervised companion objectives (UCOs). These are auxiliary objectives that – when compared to a decoder and reconstruction loss – lead to a lower OFM with the main clustering objective. The UCOs employ structure-exploiting tensor kernels, addressing the drawbacks of the naïve vectorization-based kernel for tensorial intermediate representations, such as those produced by convolutional neural networks. Our experiments with DDC and DEC show that adding the tensor kernel-based UCOs improves the clustering performance of the models, outperforming the analogous autoencoder-based approach.

## CRediT authorship contribution statement

**Daniel J. Trosten:** Conceptualization, Formal analysis, Investigation, Methodology, Project administration, Software, Validation, Visualization, Writing – original draft, Writing – review & editing. **Sigurd Løkse:** Conceptualization, Formal analysis, Investigation, Methodology, Project administration, Software, Supervision, Validation, Visualization, Writing – original draft. **Robert Jenssen:** Conceptualization, Formal analysis, Funding acquisition, Investigation, Methodology, Project administration, Software, Supervision, Validation, Visualization, Writing – original draft. **Michael Kampffmeyer:** Conceptualization, Formal analysis, Funding acquisition, Investigation, Methodology, Project administration, Software, Supervision, Validation, Visualization, Writing – original draft, Writing – review & editing.

## Declaration of competing interest

The authors declare that they have no known competing financial interests or personal relationships that could have appeared to influence the work reported in this paper.

## Data availability

Data will be made available on request.

## Acknowledgments

This work was financially supported by the Research Council of Norway (RCN), through its Centre for Research-based Innovation funding scheme (Visual Intelligence, grant no. 309439), and Consortium Partners. It was further funded by RCN FRIPRO, Norway grant no. 315029, RCN IKTPLUSS, Norway grant no. 303514, and the UiT Thematic Initiative, Norway "Data-Driven Health Technology".

**Daniel J. Trosten** received the degree of M.Sc. in machine learning and statistics in 2019 and Ph.d in 2023 from UiT the Arctic University of Norway. He is currently working as a researcher at the Earth Observation group at NORCE Norwegian Research Centre, applying and developing novel machine learning methods for remote sensing data.

**Sigurd Løkse** received the degree of M.Sc. in Electrical engineering in 2014 and Ph.D. in 2020 from UiT the Arctic University of Norway. He is currently working as a researcher at the Drones and Autonomous Systems group at NORCE Norwegian Research Centre. Research interests include information theoretic learning and learning with limited labels within deep learning.

**Robert Jenssen** directs the Visual Intelligence Centre (). He is a Professor in the Machine Learning Group () at UiT The Arctic University of Norway. He is an Adjunct Professor at the Pioneer Centre for AI, University of Copenhagen, and at the Norwegian Computing Center.

**Michael Kampffmeyer** is an Associate Professor and Head of the Machine Learning Group at UiT The Arctic University of Norway. Further, he is a Senior Researcher at the Norwegian Computing Center. Research interests include the development of deep learning algorithms that learn from limited labeled data and their interpretability.